\pdfoutput=1 %

\documentclass[11pt,a4paper]{article}
\usepackage[hyperref]{acl2019}
\usepackage{times}
\usepackage{latexsym}

\usepackage{url}

\usepackage[utf8]{inputenc}
\usepackage[T1]{fontenc} %
\usepackage{multirow}
\usepackage{booktabs}
\usepackage{amsmath}
\usepackage{amsfonts}

\usepackage{seqsplit}

\aclfinalcopy %

\title{Analyzing the Limitations of Cross-lingual Word Embedding Mappings}

\author{Aitor Ormazabal, Mikel Artetxe, Gorka Labaka, Aitor Soroa, Eneko Agirre \\
  IXA NLP Group \\
  University of the Basque Country (UPV/EHU) \\
  \texttt{aormazabal024@ikasle.ehu.eus} \\
  \texttt{\{mikel.artetxe, gorka.labaka, a.soroa, e.agirre\}@ehu.eus}}

\date{}

\begin{document}
\maketitle
\begin{abstract}
Recent research in cross-lingual word embeddings has almost exclusively focused on offline methods, which independently train word embeddings in different languages and map them to a shared space through linear transformations. While several authors have questioned the underlying isomorphism assumption, which states that word embeddings in different languages have approximately the same structure, it is not clear whether this is an inherent limitation of mapping approaches or a more general issue when learning cross-lingual embeddings. So as to answer this question, we experiment with parallel corpora, which allows us to compare offline mapping to an extension of skip-gram that jointly learns both embedding spaces. We observe that, under these ideal conditions, joint learning yields to more isomorphic embeddings, is less sensitive to hubness, and obtains stronger results in bilingual lexicon induction. We thus conclude that current mapping methods do have strong limitations, calling for further research to jointly learn cross-lingual embeddings with a weaker cross-lingual signal.

\end{abstract}

\section{Introduction}
\label{sec:introduction}

Cross-lingual word embeddings have attracted a lot of attention in recent times. Existing methods can be broadly classified into two categories: \textit{joint} methods, which simultaneously learn word representations for multiple languages on parallel corpora \citep{gouws2015bilbowa,luong2015bilingual}, and \textit{mapping} methods, which independently train word embeddings in different languages and map them to a shared space through linear transformations \cite{mikolov2013exploiting,artetxe2018generalizing}. While early work in cross-lingual word embeddings was dominated by \textit{joint} approaches, recent research has almost exclusively focused on \textit{mapping} methods, which have the advantage of requiring little or no cross-lingual signal \citep{zhang2017adversarial,conneau2018word,artetxe2018robust}.

For \textit{mapping} methods to work, it is necessary that embedding spaces in different languages have a similar structure (i.e. are approximately isomorphic), as it would otherwise be hopeless to find a linear map from one space to another. Several authors have questioned this assumption, showing that linguistic and domain divergences cause strong mismatches in embedding spaces, which in turn heavily hinders the performance of these methods \citep{sogaard2018limitations,patra2019bliss}. Nevertheless, it is not clear whether this mismatch is a consequence of separately training both embedding spaces, and thus an inherent limitation of mapping approaches, or an insurmountable obstacle that arises from the linguistic divergences across languages, and hence a more general issue when learning cross-lingual word embeddings.

The goal of this paper is to shed light on this matter so as to better understand the nature and extension of these limitations. For that purpose, we experiment with parallel corpora, which allows us to compare \textit{mapping} methods and \textit{joint} methods under the exact same conditions, and analyze the properties of the resulting embeddings. Our results show that, under these conditions, joint learning yields to more isomorphic embeddings, is less sensitive to hubness, and obtains stronger results in Bilingual Lexicon Induction (BLI). This suggests that, despite the advantage of requiring weaker cross-lingual signal, current mapping methods do have strong limitations, as they are not able to leverage the available evidence as effectively as joint methods under ideal conditions. We thus conclude that future research should try to combine the best of both worlds, exploring joint methods to learn cross-lingual word embeddings with weaker supervision.

\section{Related work}
\label{sec:related}

Cross-lingual word embeddings represent words from multiple languages in a common vector space. So as to train them, \textbf{joint methods} simultaneously learn the embeddings in the different languages, which requires some form of cross-lingual supervision. This supervision usually comes from parallel corpora, which can be aligned at the word level \citep{luong2015bilingual}, or only at the sentence level \citep{gouws2015bilbowa}. In addition to that, methods that rely on comparable corpora \citep{vulic2016bilingual} or large bilingual dictionaries \citep{duong2016learning} have also been proposed. For a more detailed survey, the reader is referred to \citet{ruder2017survey}.

In contrast, \textbf{offline mapping} approaches work by aligning separately trained word embeddings in different languages. For that purpose, early methods required a training dictionary, which was used to learn a linear transformation that mapped these embeddings into a common space \citep{mikolov2013exploiting,artetxe2018generalizing}. The amount of required supervision was later reduced through self-learning methods \citep{artetxe2017learning}, and then completely eliminated through adversarial training \citep{zhang2017adversarial,conneau2018word} or more robust iterative approaches combined with initialization heuristics \citep{artetxe2018robust,hoshen2018nonadversarial}.

There are several authors that have discussed the potential \textbf{limitations} of these mapping approaches. For instance, \citet{sogaard2018limitations} observe that the assumption that separately trained embeddings are approximately isomorphic is not true in general, showing that the performance of mapping methods is conditioned by the language pair, the comparability of the training corpora, and the parameters of the word embedding algorithms. Similarly, \citet{patra2019bliss} show that the isomorphism assumption weakens as the languages involved become increasingly etymologically distant.
Finally, \citet{nakashole2018characterizing} argue that embedding spaces in different languages are linearly equivalent only at local regions, but their global structure is different.
Nevertheless, neither of these works does systematically analyze the extent to which these limitations are inherent to mapping approaches. To the best of our knowledge, ours is the first work comparing joint and mapping methods in the exact same conditions, characterizing the nature and impact of such limitations.

\section{Experimental design}
\label{sec:design}

We next describe the cross-lingual embedding methods, evaluation measures and datasets used in our experiments.

\subsection{Cross-lingual embedding methods}
\label{sec:build-biling-embedd}

We use the following procedure to learn cross-lingual embeddings, which are representative of the state-of-the-art in mapping and joint methods:

\noindent \textbf{Mapping:} We first train 300-dimensional skip-gram embeddings for each language using \emph{word2vec} \citep{mikolov2013distributed} with 10 negative samples, a sub-sampling threshold of 1e-5 and 5 training iterations. Having done that, we map the resulting monolingual embeddings to a cross-lingual space using the unsupervised mode in VecMap\footnote{\url{https://github.com/artetxem/vecmap}} \citep{artetxe2018robust}, which builds an initial solution based on heuristics and iteratively improves it through self-learning.

\noindent \textbf{Joint learning:} We use the BiVec\footnote{\url{https://github.com/lmthang/bivec}} tool proposed by \citet{luong2015bilingual}, an extension of skip-gram that, given a word aligned parallel corpus, learns to predict the context of both the source word and the target word aligned with it. For that purpose, we first word align our training corpus using FastText \citep{dyer2013simple}. Given that BiVec is a natural extension of skip-gram, we use the exact same hyperparameters as for the mapping method.

In both cases, we restrict the vocabulary to the most frequent 200,000 words.

\begin{table*}[t]
\begin{center}

\begin{small}
  \begin{tabular}{llccccccccccccc}
    \toprule
    && \multirow{2}{*}{Eig.} && \multicolumn{2}{c}{Hub. NN ($\uparrow$)} && \multicolumn{2}{c}{Hub. CSLS ($\uparrow$)} && \multicolumn{2}{c}{P@1 Eparl ($\uparrow$)} && \multicolumn{2}{c}{P@1 MUSE ($\uparrow$)} \\
    \cmidrule{5-6} \cmidrule{8-9} \cmidrule{11-12} \cmidrule{14-15}
    && sim. ($\downarrow$) && 10\% & 100\% && 10\% & 100\% && NN & CSLS && NN & CSLS \\
    \midrule                   
    \multirow{2}{*}{FI-EN}
    & Joint learning & \bf 28.9  && \bf 0.45 & \bf 52.8 && \bf 1.13 & \bf 57.5 && \bf 65.2 & \bf 68.3 && \bf 83.4 & \bf 85.2 \\
    & Mapping        & 115.9 && 0.12 & 33.8 && 0.38 & 46.1 && 26.3 & 34.8 && 44.6 & 56.8 \\
    \midrule               
    \multirow{2}{*}{ES-EN}
    & Joint learning &  \bf 31.2 && \bf 0.65 & \bf 66.0 && \bf 1.40 & \bf 71.3 && \bf 68.7 & \bf 69.3 && \bf 91.9 & \bf 92.4 \\
    & Mapping        &  47.8 && 0.58 & 63.1 && 1.31 & 69.1 && 65.4 & 67.0 && 87.1 & 89.0 \\
        \midrule               
    \multirow{2}{*}{DE-EN}
    & Joint learning &  \bf 32.8 && 0.58 & \bf 58.8 && 1.29 & \bf 65.2 && \bf 70.6 & \bf 70.4 && \bf 90.1 & \bf 89.2 \\
    & Mapping        &  39.4 && \bf 0.60 & 58.7 && \bf 1.33 & 64.8 && 65.3 & 66.4 && 82.4 & 83.1 \\
        \midrule            
    \multirow{2}{*}{IT-EN}
    & Joint learning &  \bf 26.5 && \bf 0.75 & \bf 69.7 && \bf 1.61 & \bf 74.2 && \bf 71.5 & \bf 71.8 && \bf 90.6 & \bf 90.0 \\
    & Mapping        &  43.9 && 0.65 & 63.9 && 1.53 & 70.8 && 64.1 & 67.2 && 84.4 & 85.9 \\
    \bottomrule
  \end{tabular}
\end{small}
\end{center}
\caption{Evaluation measures for the two cross-lingual embedding approaches. Arrows indicate whether lower ($\downarrow$) or higher ($\uparrow$) is better. See text for further details.}
\label{tab:results}
\end{table*}

\subsection{Evaluation measures}
\label{sec:metrics}

We use the following measures to characterize cross-lingual embeddings:

\noindent \textbf{Isomorphism}. %
Intuitively, the notion of isomorphism captures the idea of how well the embeddings in both languages fit together (i.e. the degree of their structural similarity).
So as to measure it, we use the eigenvalue similarity metric proposed by \citet{sogaard2018limitations}. For that purpose, we first center and normalize the embeddings, calculate the nearest neighbor graphs of the $10,000$ most frequent words in each language, and compute their Laplacian matrices $L_1$ and $L_2$. We then find the smallest $k_1$ such that the sum of the largest $k_1$ eigenvalues of $L_1$ is at least $90\%$ of the sum of all its eigenvalues, and analogously for $k_2$ and $L_2$. Finally we set $k=min(k_1,k_2)$, and define the eigenvalue similarity of the two spaces as the sum of the squared differences between the $k$ largest eigenvalues of $L_1$ and $L_2$, $\Delta = \sum_{i=1}^k (\lambda_{1_i} - \lambda_{2_i})^2$. %

\noindent \textbf{Hubness}. Cross-lingual word embeddings are known to suffer from the hubness problem \citep{radovanovic2010hubs,radovanovic2010existence,dinu2015improving}, which causes a few points (known as \textit{hubs}) to be the nearest neighbors of many other points in high-dimensional spaces. So as to quantify it, we measure the minimum percentage of target words $H_{N}$ that are the nearest neighbor of at least $N\%$ of the source words, where $N$ is a parameter of the metric.\footnote{Some previous work uses an alternative hubness metric that computes the hubness level $N(t)$ of each target word $t$ (i.e. the number of source words whose nearest neighbor is $t$) and measures the skewness of the resulting distribution. However, we find this metric to have two important drawbacks: 1) its magnitude is not easily interpretable, and 2) it is invariant to the variance of the distribution, even if higher variances are indicative of a higher hubness level. For instance, we observed that two very similar spaces (produced running word2vec twice over the same corpora) mapped to each other produced unusually high skewness scores, caused by the scale normalization done in skewness (division by the standard deviation).}
For instance, a hubness value of $H_{10\%}=0.3\%$ would indicate that 0.3\% of the target words are the nearest neighbors of 10\% of the source words. This way, lower values of $H_N$ are indicative of a higher level of hubness, and the parameter $N$ serves to get a more complete picture of the distribution of hubs. For brevity, we report results for $N=10\%$ and $100\%$.
While the nearest neighbor retrieval is usually done according to cosine similarity, \citet{conneau2018word} proposed an alternative measure, called Cross-domain Similarity Local Scaling (CSLS), that penalizes the similarity scores of hubs, which in turn reduces the hubness level. So as to better understand its effect, we report results for both CSLS and standard nearest neighbor with cosine similarity (NN).

\noindent \textbf{Bilingual Lexicon Induction (BLI)}. Following common practice, we induce a bilingual dictionary by linking each word in the source language with its nearest neighbor in the target language. So as to evaluate the quality of the induced translations, we compare them to a gold standard dictionary, and measure the precision at 1. We report results for both nearest neighbor with cosine similarity (NN) and the aforementioned CSLS retrieval. Note that, in addition to having a practical application, BLI performance is an informative measure of the quality of the embeddings, as a good cross-lingual representation should place equivalent words close to each other.

\subsection{Datasets}
\label{sec:resources}

We experiment with 4 language pairs with English as the target language, covering 3 relatively close languages (German, Spanish and Italian) and a non-indoeuropean agglutinative language (Finnish). All embeddings were trained on the BiCleaner v3.0 version of the ParaCrawl corpus,\footnote{\url{https://paracrawl.eu/}} a parallel corpus collected through crawling and filtered according to \citet{prompsit:2018:WMT}.
The size of this corpus changes from one language to another: German and Spanish are the largest (503 and 492 million tokens in the English side, respectively), followed by Italian (308 million tokens), and Finnish (55 million tokens). %

As for the evaluation dictionaries for BLI, we use two datasets that have been widely used in the literature. The first one, which we call Eparl, was first introduced by \citet{dinu2015improving} and subsequently extended by \citet{artetxe2017learning} and \citet{artetxe2018generalizing}, and consists of 1,500 test entries extracted from Europarl word alignments and uniformly distributed in 5 frequency bins. The second one, which we call MUSE, consists of another 1,500 test entries, and was compiled by \citet{conneau2018word} using internal translation tools.

\section{Results}
\label{sec:results}

Table \ref{tab:results} reports the results of all the evaluation measures for both cross-lingual embedding approaches.

The eigenvalue similarity metric shows that joint learning obtains substantially more \textbf{isomorphic} embedding spaces than the mapping approach, indicating that the representations it learns for different languages have a more similar structure. At the same time, it is remarkable that the eigenvalue similarity for the four language pairs is very close in the case of joint learning, with values that range between 26.5 and 32.8. In contrast, the degree of isomorphism for Finnish-English is substantially lower than the rest in the case of the mapping approach, which is likely caused by the typological differences between these languages and the smaller size of the training corpus. This suggests that joint learning is able to appropriately fit divergent languages together, which is troublesome when the embedding spaces are learned separately and then mapped together.

When it comes to \textbf{hubness}, our results show that joint learning is generally less sensitive to this problem, although differences greatly vary depending on the language pair. This way, both approaches have a similar behavior in German, while joint learning does moderately better for Spanish and Italian, and the difference becomes very large for Finnish. Once again, this suggests that mapping methods are more severely affected by linguistic divergences. At the same time, we observe that CSLS is very effective at reducing the hubness level, specially for offline mapping.

Finally, we observe that joint learning outperforms offline mapping in \textbf{BLI}. This difference is particularly pronounced for Finnish-English (e.g. 26.3\% vs 65.2\% for NN on Eparl), which is in line with the general behavior observed so far. At the same time, our results show that CSLS is most helpful with offline mapping, but it even has a negative impact with joint learning for some language pairs. This can be partly explained by the fact that the latter approach is less sensitive to hubness, which CSLS tries to address.

\section{Discussion}
\label{sec:discussion}

Our analysis reveals that, when trained on parallel corpora under the exact same conditions, joint learning obtains substantially better cross-lingual representations than offline mapping, yielding to more isomorphic embeddings that are less sensitive to hubness and obtain stronger results on BLI. Moreover, our results show that divergences across languages can be effectively mitigated by jointly learning their representations, whereas trying to align separately trained embeddings is troublesome when such divergences exist.

Note that this should not be interpreted as a claim that existing joint methods are superior to existing mapping methods. In fact, we believe that both families serve different purposes in that they require a different degree of supervision (e.g. mapping methods can exploit monolingual corpora, which is useful in practical settings), so the choice to use one approach or the other should depend on the resources that are available in each particular case. Nevertheless, our results do show that offline mapping has fundamental limitations that, given the available evidence, seem specific to this particular approach.

For that reason, we argue that, while recent research on cross-lingual word embeddings has almost exclusively focused on mapping methods, future work should consider alternative approaches to try to overcome the limitations of this paradigm. In particular, we believe that an interesting direction would be to adapt joint methods so they can work with monolingual corpora.

\section{Conclusions and future work}
\label{sec:conclusions}

In this work, we compare the properties of cross-lingual word embeddings trained through joint learning and offline mapping on parallel corpora. We observe that, under these ideal conditions, joint learning yields to more isomorphic embeddings, is less sensitive to hubness, and obtains stronger results in bilingual lexicon induction, concluding that current mapping methods have strong limitations.

This analysis calls for further research on alternatives to current mapping methods, which have been very successful on unsupervised settings. In particular, we would like to explore new methods to jointly learn cross-lingual embeddings on monolingual corpora.

\section*{Acknowledgments}

This research was partially supported by the Spanish MINECO (UnsupNMT TIN2017‐91692‐EXP and DOMINO PGC2018-102041-B-I00, cofunded by EU FEDER), the BigKnowledge project (BBVA foundation grant 2018), the UPV/EHU (excellence research group), and the NVIDIA GPU grant program. Mikel Artetxe was supported by a doctoral grant from the Spanish MECD.

\bibliography{acl2019}
\bibliographystyle{acl_natbib}

\end{document}